\documentclass[runningheads]{llncs}
\usepackage[T1]{fontenc}
\usepackage[hyperindex,linktocpage, colorlinks=true, urlcolor=blue, linkcolor=blue, citecolor=blue]{hyperref}  

\usepackage{graphicx}
\usepackage{amsmath} 
\usepackage{amssymb}
\usepackage{orcidlink}
\usepackage{multirow}
\begin{document}
\title{LLM-Assisted Automated Deductive Coding of Dialogue Data: Leveraging Dialogue-Specific Characteristics to Enhance Contextual Understanding}
\titlerunning{LLM-Assisted Automated Deductive Coding of Dialogue Data}

\author{Ying Na \and Shihui Feng\orcidlink{0000-0002-5572-276X}}
\authorrunning{Y. Na and S. Feng}

\institute{The University of Hong Kong, Hong Kong SAR, China\\
\email{\{naying,shihuife\}@hku.hk}}

\maketitle            
\begin{abstract}
Dialogue data has been a key source for understanding learning processes, offering critical insights into how students engage in collaborative discussions and how these interactions shape their knowledge construction. The advent of Large Language Models (LLMs) has introduced promising opportunities for advancing qualitative research, particularly in the automated coding of dialogue data. However, the inherent contextual complexity of dialogue presents unique challenges for these models, especially in understanding and interpreting complex contextual information. This study addresses these challenges by developing a novel LLM-assisted automated coding approach for dialogue data. The novelty of our proposed framework is threefold: 1) We predict the code for an utterance based on dialogue-specific characteristics---communicative acts and communicative events---using separate prompts following the role prompts and chain-of-thoughts methods; 2) We engaged multiple LLMs including GPT-4-turbo, GPT-4o, DeepSeek in collaborative code prediction; 3) We leveraged the interrelation between events and acts to implement consistency checking using GPT-4o. In particular, our contextual consistency checking provided a substantial accuracy improvement. We also found the accuracy of act predictions was consistently higher than that of event predictions. This study contributes a new methodological framework for enhancing the precision of automated coding of dialogue data as well as offers a scalable solution for addressing the contextual challenges inherent in dialogue analysis.

\end{abstract}
\section{Introduction}
In recent years, large language models (LLMs) have gained considerable attention for their potential applications in qualitative research, particularly in deductive coding \cite{1-xiao2023supporting}. With their advanced text-processing capabilities, LLMs hold significant promise in automating and enhancing the analysis of complex text data. However, a critical limitation persists: the inability of LLMs to consistently understand and interpret context remains a major challenge \cite{3-zhang2024qualitative}. Research has shown that LLMs often exhibit issues such as inconsistency \cite{3-zhang2024qualitative} and contextual forgetfulness \cite{4-zhai2023investigating}, where they fail to maintain a coherent understanding of the data over time. This limitation becomes particularly pronounced and may introduce additional biases when analyzing dialogue data, such as interviews, which are inherently complex, context-dependent, and characterized by dynamic, interactive exchanges \cite{5-ashwin2023using}.

Dialogue, as a dynamic process, involves multiple layers of interaction and meaning-making, and its complexity is addressed through various analytical frameworks \cite{32-song2020systematic}.Understanding dialogue data is crucial for gaining insights into student engagement in learning processes, particularly in collaborative learning contexts \cite{6-chen2023linkages}.  Within the Ethnography of Communication, dialogue is categorized into distinct communicative components: Communicative Acts (CAs), Communicative Events (CEs), and Communicative Situations (CSs) \cite{9-saville2008ethnography}. Communicative Acts, such as questioning or responding, are the fundamental units of dialogue. These acts combine to form Communicative Events, which encompass coherent sequences involving participants, purposes, tasks, and themes. At the highest level, Communicative Situations provide the broader contextual and structural framework within which dialogue unfolds \cite{9-saville2008ethnography}.  In dialogic teaching and learning environments, dialogue reflects this hierarchical structure, comprising multiple layers of interaction—each containing distinct events, and each event comprising various acts \cite{10-hennessy2016developing}. From the individual acts to the overarching situational context, each layer shapes the communication dynamics, influencing how meaning is constructed and exchanged throughout the learning process~\cite{9-saville2008ethnography}. Effectively analyzing dialogue requires capturing these interconnected dimensions, which poses a significant challenge for large language models (LLMs).

Currently, many coding frameworks either merge events and acts into a single code or only focus on events. In our study, we aim to address this limitation by separating events and acts as independent communicative components for prediction within each dialogue interaction. We can leverage the inherent complexity of dialogue and the interdependencies between different communicative components to assist LLMs in understanding the intricate nature of dialogue data, thereby reducing their cognitive burden.  This study seeks to bridge these gaps by developing an LLM-assisted coding approach specifically designed for dialogue data. By harnessing advanced LLMs and capturing the multi-dimensional and dynamic characteristics of dialogue, we aim to improve coding accuracy and address the challenges LLMs encounter in interpreting complex dialogue contexts.  Our primary research question is: How can LLM-assisted methods address the challenges of understanding complex dialogue data contexts and overcome LLMs’ limitations in maintaining contextual consistency?

\section{Related Work}

In recent years, qualitative coding based on LLMs has yielded valuable findings and methodological insights, including various research frameworks that suggest LLMs hold potential for streamlining an automated qualitative data analysis \cite{12-dai2023llm}. Despite this promise, their application in the educational field remains limited, with human feedback still playing a critical role in the coding process \cite{12-dai2023llm}. Comparative studies have revealed that hybrid methods, which integrate LLMs at different stages of codebook development, tend to outperform fully automated approaches, as sole reliance on models like ChatGPT often fails to produce satisfactory results \cite{14-barany2024chatgpt}.

While LLMs have demonstrated potential in dialogue-related tasks, such as multi-turn dialogue understanding and spoken language comprehension, limitations remain \cite{15-pan2023preliminary}. Studies have indicated that although LLM-based coding techniques can achieve fair to substantial agreement with human raters in context-independent tasks, they face challenges in context-dependent coding dimensions \cite{16-hou2024prompt}. Autonomous coding of interview data by models like ChatGPT and Llama-2 has further highlighted the difficulty of adapting to nuanced, context-specific tasks, which are essential for qualitative research due to the intricate and flexible interpretations required \cite{5-ashwin2023using}.

To address these limitations, recent advancements have focused on leveraging contextual knowledge more effectively \cite{17-liu2024bridging}. Approaches integrating retrieval-augmented generation (RAG) workflows have demonstrated superior performance by enabling LLMs to handle contextual information more efficiently \cite{19-zhang2024automatic}. Furthermore, research has underscored the importance of combining LLMs with traditional natural language processing (NLP) techniques to enhance coding at multiple levels, such as sentence, slot, and semantic coding \cite{20-bryda2024words}. These studies emphasize that LLMs must go beyond simple keyword recognition to capture deeper linguistic elements, including assumptions, values, metaphors, and power dynamics \cite{20-bryda2024words}.

However, many existing approaches to coding dialogue data rely on prediction methods that are not specifically tailored to dialogue’s unique attributes. This lack of specialization limits their ability to consistently address contextual shifts and maintain coherence in analysis. Manual coding, while effective, is resource-intensive and impractical for large-scale datasets. As a result, there is a need for more efficient methods that address both the contextual intricacies of dialogue and the limitations of LLMs.  

\section{Methodology}

\subsection{Learning Context and Data Collection 
}

In this study, data was collected from student discussions during a collaborative problem-solving task conducted in a laboratory setting. 24 participants worked in 8 groups with three students in each group to complete a set of multiple choice questions and a short essay related to two education theories: Bloom's Taxonomy \cite{30-forehand2010bloom} and the ICAP Framework \cite{31-chi2014icap}. Each group was tasked with completing a set of multiple-choice questions, which assessed their understanding of the components of the two theories. In addition, each group was required to write a short essay exploring the application and understanding of these two theories within practical teaching scenarios. The participants were given one hour to complete the tasks, during which they were encouraged to engage in open discussions and utilize various information sources to explore the content. The entire discussion process was recorded with the participants' consent. All user data were strictly anonymized throughout the analysis, with no user information incorporated into the prompts.

\subsection{Coding Framework}\label{cf}

Due to the characteristics of dialogue data, we decompose its communicative components by analyzing it from three perspectives: Interaction (CS), Event (CE), and Act (CA) \cite{9-saville2008ethnography}. Based on our dataset, we coded communicative situations into four types---cognitive interactions, metacognitive interactions, socio-emotional interactions, and coordinative interactions---each representing a distinct dimension of collaborative learning. Cognitive interactions focus on the exchange, generation, and application of knowledge and ideas \cite{21-king2007scripting}. Metacognitive interactions involve the regulation and monitoring of cognitive processes to enhance group efficiency, such as planning strategies or monitoring and evaluating progress \cite{22-akturk2011literature}. Socio-emotional interactions address the emotional and interpersonal dynamics within the group \cite{37-socioemointer}. Lastly, coordinative interactions are concerned with organizing group activities to ensure smooth collaboration. Under the four types of interactions, there are various events (refer to Table~\ref{table-interactions}). These events represent the ones captured from our data.

\begin{table}[h]
\centering
\caption{\centering Layered Coding Framework of Group Interactions in Collaborative Problem Solving}
\label{table-interactions}
\begin{tabular}{@{}p{2cm}|p{2.2cm}|p{4cm}|p{4cm}@{}}
\hline
Interactions & Event & Definition & Example \\
\hline
Cognitive Interactions & Concept Exploration & Discuss or clarify concepts related to learning tasks. & \textit{\textquotedblleft What is the bloom taxonomy?\textquotedblright} \\
\cline{2-4}
& Solution Development & Discuss or clarify solutions for learning tasks. & \textit{\textquotedblleft The answer should be the last one.\textquotedblright} \\
\hline
Metacognitive Interactions & Planning & Plan problem-solving procedures, including goal setting. & \textit{\textquotedblleft We can first search for the answer.\textquotedblright} \\
\cline{2-4}
& Evaluating & Assess information, and task outcomes. & \textit{\textquotedblleft GPT results lack detail.\textquotedblright} \\
\cline{2-4}
& Monitoring & Evaluate task progress and ensure alignment with plans. & \textit{\textquotedblleft We can move to the next question.\textquotedblright} \\
\hline
Coordinative Interactions & Coordinate Participants & Allocate tasks and assign roles within the team. & \textit{\textquotedblleft We can divide the task into three parts.\textquotedblright} \\
\cline{2-4}
& Coordinate Procedures & Organize workflow and resolve technical issues. & \textit{\textquotedblleft You go first.\textquotedblright} \\
\hline
Socio-Emotional& Emotional Expression & Express feelings about group work. & \textit{\textquotedblleft That's hilarious!\textquotedblright} \\
\cline{2-4}Interactions& Encouragement & Praise or cheer up others. & \textit{\textquotedblleft Thank you!\textquotedblright} \\
\cline{2-4}
& Self-disclosure & Discuss personal issues or task unfamiliarity. & \textit{\textquotedblleft I’ve worked as a TA before.\textquotedblright} \\
\hline
\end{tabular}
\end{table}

Each interaction not only conveys an event element but also expresses an underlying behavioral intent, which is context-dependent and often appears in paired forms (such as ``ask and answer''). The relationship between events and acts can be understood as a hierarchical structure. For example, ``concept exploration-ask'' represents an ``ask act'' occurring within the ``concept exploration event''. Each act is inherently tied to a specific event, serving as its functional or communicative unit. A sequence of related acts collectively constitutes an event, where the acts define the specific behaviors taking place within the broader context of the event. In this way, events provide the overarching framework, while acts represent the granular actions that drive and shape the event.

Our study identifies six key acts for each event, all of which are frequently observed in our data. Each act has a distinct meaning: ``Ask'' refers to asking relevant questions about the specific content, ``Answer'' refers to providing responses, and these two typically occur in pairs. ``Give'' refers to expressing thoughts, suggestions, or statements about the content without being prompted by a question, while ``Agree'' indicates agreement without further elaboration. ``Build on'' refers to an agreement with additional explanation, and ``Disagree'' signifies disagreement. Socio-emotional interactions were not included in the framework, as these interactions are often isolated expressions, such as "lol." If future research identifies acts within socio-emotional interactions, this part of the framework can be expanded accordingly.

Based on the layered contextual framework, two researchers coded validation and test sets with ground truth code labels to be used for the prediction process. The inter-annotator agreement between the two researchers was 96.11\% overall, assessed using Cohen's Kappa statistic.

\subsection{LLM-based Coding Procedures}

The total number of utterances collected is 5676. To ensure both efficiency and resource optimization in the prediction process, we randomly selected a subset of 30\% of the data as the validation set for calibration. This portion of the data was manually coded to generate a ground truth, serving as the target for model adjustment. On the validation set, we utilized in-context learning (ICL) to incorporate human knowledge into large language models (LLMs) by modifying the prompts\cite{34-dong2024surveyincontextlearning}. We iterated through different coding strategies and configurations, testing various models and prediction methods to determine which was most effective. During each round of optimization, results were evaluated using Inter-Annotator Agreement (IAA) \cite{23-landis1977measurement}. We used Cohen’s Kappa statistic to assess the consistency across different rounds of model-generated results. Once we achieved an IAA score of 80\% between the machine-generated and human-generated coding results, we considered this prediction scheme reliable enough for broader application, following \cite{23-landis1977measurement}, and completed the validation process. This step-by-step refinement ensured that the final model was both accurate and resource-efficient, without compromising the quality of the predictions. 

\begin{figure}
\centering
\includegraphics[width=0.9\textwidth]{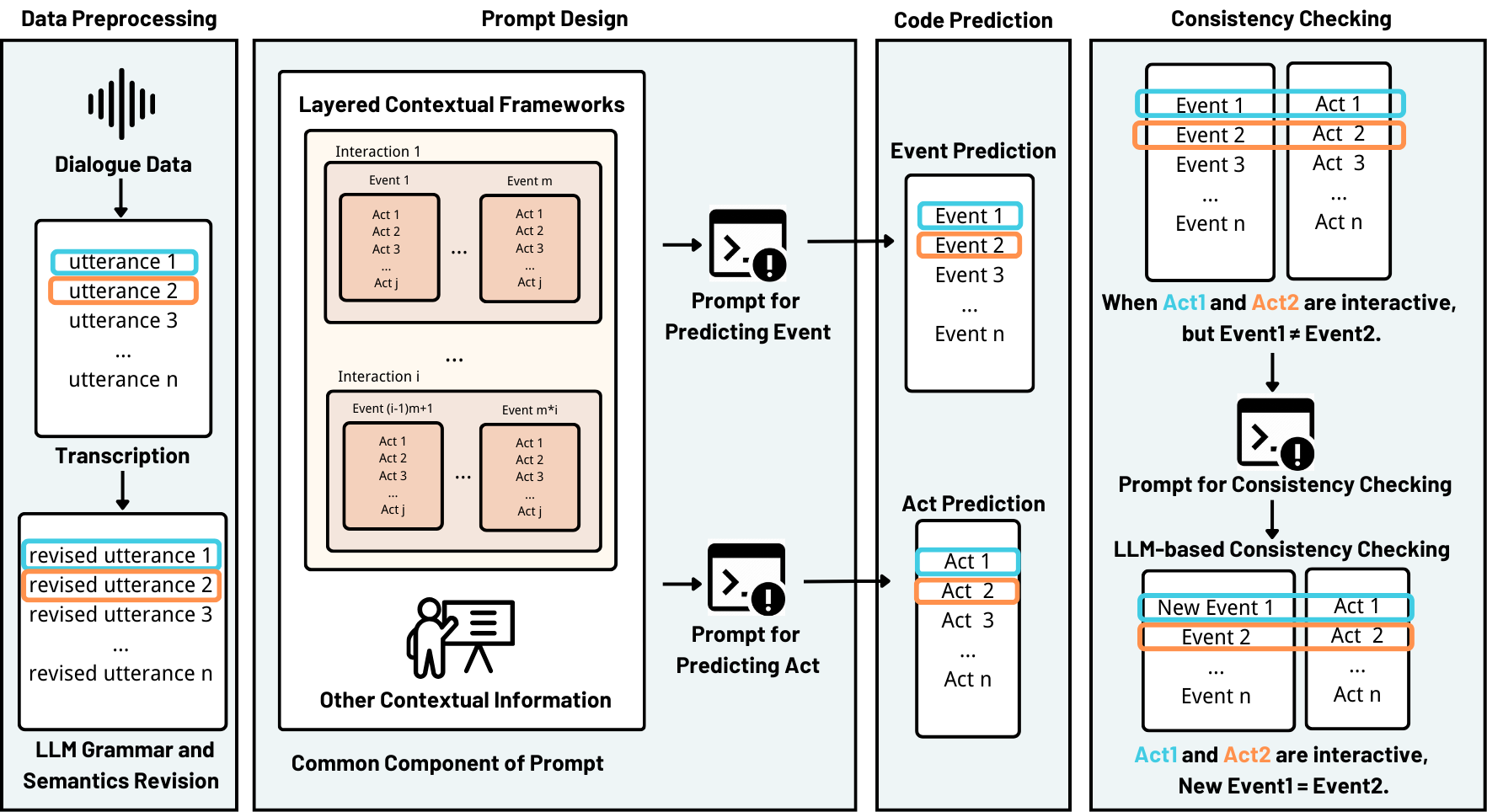}

\caption{\centering The LLM-Assisted Contextually Consistent Dialogue Coding Framework
}\label{fig-framework}
\end{figure}

Before deploying the model to the full dataset, we randomly selected 10\% of the data as a test set to evaluate the model's robustness and to assess whether the prompt, as a critical "parameter," could be generalized to a broader context, considering that prompts may perform differently across datasets. This portion was also manually coded for ground truth code labels, which were then compared with the LLM predictions using the validated model. Our model maintained a consistency of greater than 80\% on the test set, allowing us to proceed to expand its application to the remaining 60\% of the dataset. In the following sections we will describe the proposed LLM-assisted dialogue coding framework(refer to Fig.~\ref{fig-framework}), which was evaluated using the validation and test procedures discussed above.

\vspace{-2em}
\subsubsection{Data Preprocessing}
The audio recordings were transcribed into text using Whisper, which is suited for accurately processing multi-speaker dialogue data. Each individual's utterance is treated as the unit of coding. Whisper captured the spoken words and comprehensive metadata, including fields such as “speaker,” “text,” “start,” and “end.”  The transcription process resulted in a dataset containing 5676 structured text entries, each representing a discrete conversational segment. 

\begin{figure}
\centering
\includegraphics[width=0.9\textwidth]{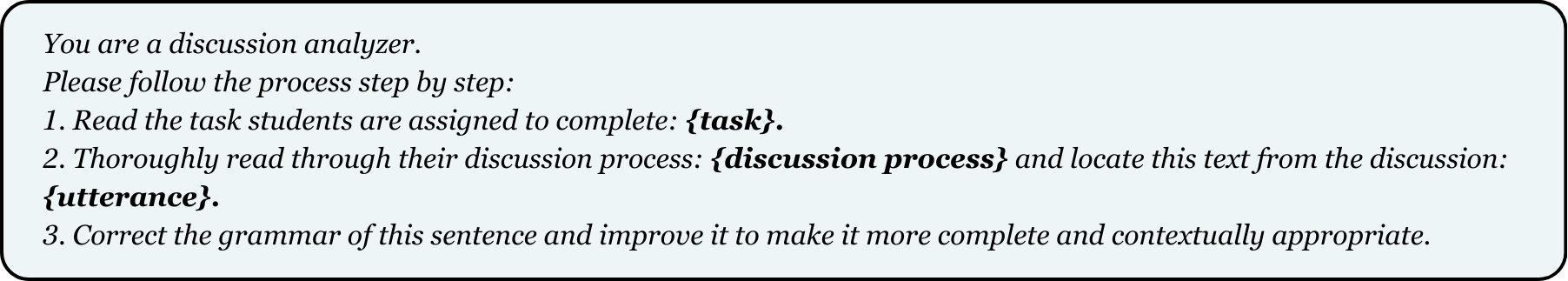}
\caption{\centering Prompt for Grammar and Semantics Revision
}\label{fig-promptGSR}
\end{figure}

After Whisper transcription, another large language model (LLM) was employed to refine both the grammar and semantics of the transcribed text (refer to Fig. ~\ref{fig-promptGSR}). Understanding individual utterances in dialogue often requires integrating complex contextual cues, including physical gestures, body language and other shared contextual information \cite{25-mlakar2023understanding}. However, audio data frequently lacks these critical elements, and transcription processes also introduce some inaccuracies or omit unintelligible segments. For instance, in our raw data a participant simply uttered the word “validate” without further explanation, which could easily be misconstrued. In the actual discussion, the term referred  to the process of selecting an answer to a multiple-choice question after reviewing retrieved content. However, without adequate context, such isolated terms are prone to misinterpretation, leading to inaccurate coding. We used GPT-4o to optimize the grammar and pragmatics of the text while supplementing its semantics and embedding the necessary task content and discussion context directly into the sentences requiring coding. This approach ensured that each sentence contained enough context to support accurate interpretation. For instance, the previously ambiguous term “validate” was transformed into a contextually enriched statement: “I think validation is about applying" (the correct interpretation in the context of Bloom's taxonomy).  

\subsubsection{Prompt Design}
The design of effective prompts is a critical step in ensuring that LLMs can accurately understand and perform coding tasks. In this study, we adopted two main strategies: role prompts \cite{27-kong2023better} and chain-of-thought methods \cite{28-wei2022chain}. These strategies help guide the models in understanding the context and the rules of the coding task, leading to more accurate predictions. A role prompt clearly defines the model's role, guiding its reasoning process in a specific context \cite{27-kong2023better}. The chain-of-thought method encourages the model to engage in step-by-step reasoning, which clarifies how to derive the appropriate coding labels from the given information \cite{28-wei2022chain}. We also provide the model with the layered contextual framework, entire discussion process, and the utterance to be coded as part of the prompt. 

\begin{figure}
\centering
\includegraphics[width=0.9\textwidth]{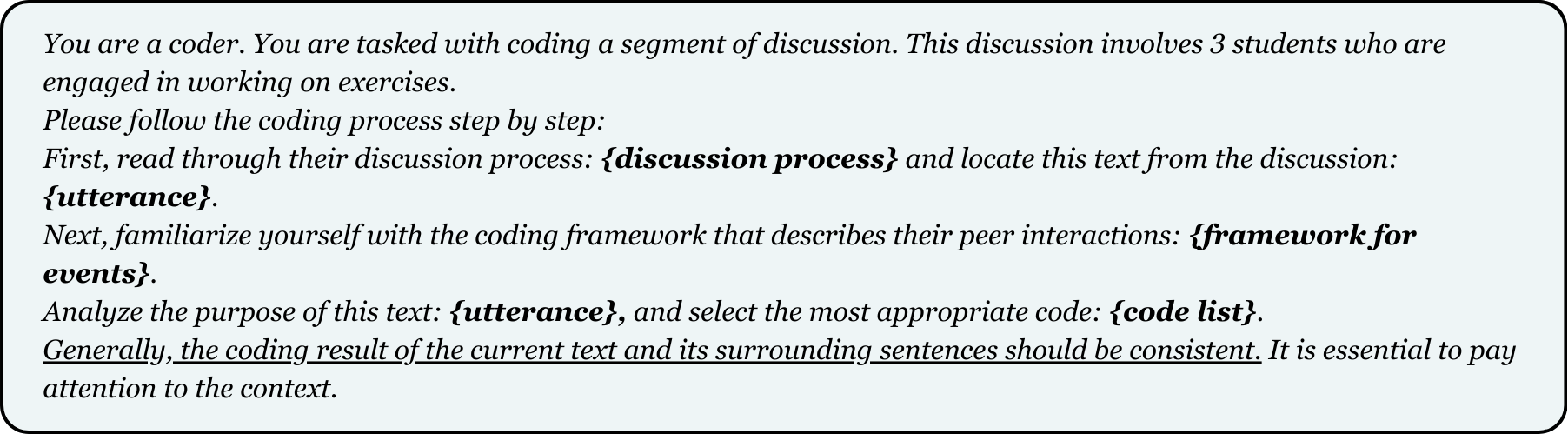}
\caption{\centering Prompt for Predicting Event}\label{fig-event}
\end{figure}

\begin{figure}
\centering
\includegraphics[width=0.9\textwidth]{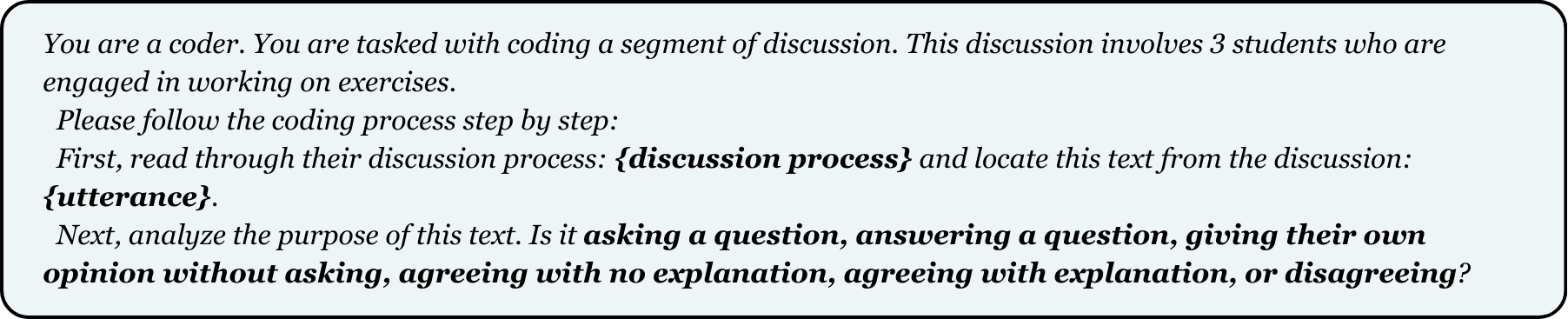}
\caption{\centering Prompt for Predicting Act}\label{fig-act}
\end{figure}

We considered two approaches when designing the prompts. The first option was to have the LLM predict both the event and the act simultaneously, such as ``solution development-ask.'' The second option involved dividing the prediction process into two distinct parts: first predicting the event, and then predicting the act. After comparing the results of the two prediction approaches, we found that separating the predictions for event and act yielded comparatively higher Cohen’s kappa values (refer to Table~\ref{fig-sc}). Below are the separate instructions for the prediction of events and acts (Fig.~\ref{fig-event} and Fig.~\ref{fig-act} respectviely). In the event prediction prompt, we emphasize the importance of ensuring consistency with the surrounding context.


\subsubsection{Code Prediction}
During the prediction process, we explored various strategies aimed at enhancing the accuracy and consistency of the results. Initially, we utilized individual large language models (LLMs), including GPT-4-turbo, GPT-4o, and DeepSeek-chat to generate predictions and compare their outputs. To ensure high consistency and accuracy in their predictions, we carefully adjusted their initialization settings. However, the models continued to display inconsistencies and failed to meet the desired performance benchmarks (refer to Table~\ref{fig-sc}). 

To overcome the limitations of individual model predictions, we treated the coding process as a collaborative decision-making problem, combining the predictions of multiple models to make the final decision. Specifically, we used three different large language models(GPT-4-turbo, GPT-4o, and DeepSeek-chat) as contributors. Each model provided $k$ predictions, resulting in a total of $3k$ predictions per coding task. The final prediction was chosen as the one with the highest weighted frequency. If two or more predictions had the same frequency for a label prediction, the models made additional predictions for the label until a majority was obtained.

For a given code label prediction, we calculate the weighted frequency $F_c$ of each label candidate $c$ across all LLMs as
\begin{equation}\label{eq:freq}
F_c = \sum_{i=1}^{z} \sum_{j=1}^{k} w_i \delta(P_{ij},c),
\end{equation}
where $P_{ij}$ is the $j$-th predicted class label from model $i$, and $w_i$ are weights assigned to each model $i$ (set equal at $w_i=1$, $z=3$, $k=5$ for this task). 

The indicator function $\delta(P_{ij}, c)$ is defined as:
\[
\delta(P_{ij}, c) =
\begin{cases}
1 & \text{if } P_{ij} = c, \\
0 & \text{if } P_{ij} \neq c.
\end{cases}
\]

The final coding prediction $c_{\text{final}}$ is then given by
\begin{equation}\label{eq:cfinal}
 c_{\text{final}} = \arg max_{c \in C} F_c,
\end{equation}
where $C$ is the set of codes predicted for the given input.

\subsubsection{Consistency Checking}
During the prediction process, we separated events and acts for analysis. However, as interrelated components, they must be linked to construct the various interactions that constitute communication. The event is a higher-level component of the act, serving as the context for the act. The basic consensus is that consecutive and interactive acts should be within the same event \cite{10-hennessy2016developing}. For example, in acts involving questioning and answering, the event must be consistent to ensure that the corresponding codes remain aligned. This consistency and relevance are essential for accurately reflecting the dynamics of the conversation. Since the accuracy of act prediction is higher than that of event prediction, we use the act as a reference base to assist in modifying the event prediction results. Based on the Ethnography of Communication, a communicative sequence, as defined within the act sequence component, describes the ordering of communicative acts within the same event \cite{9-saville2008ethnography}. In our study, based on the interactive patterns of turn-taking and idea-building in group discussions\cite{35-SKANTZE2021101178}, we identified several pairs of communicative sequence, such as ask/answer, give/agree, give/disagree, and give/build on. When those communicative sequences have different events, we prompt the LLM to refer to the overall context and modify the code accordingly.

We employed GPT-4o to perform consistency checks(CC). GPT-4o was provided with a layered coding framework and contextual information, including the current sentence and its corresponding predicted code, as well as the subsequent sentences and their respective predicted codes. Based on this analysis, it evaluated whether the event in the current code was consistent with the event code of the next sentence, and whether the act in the current code was interactive with the act of the next sentence. If the current event was found to be inconsistent with the next event code during interactive acts, the prompted model referred to the context to determine which event was correct, which required modification, and proposed a revised code for the current prediction. Starting from the first sentence, we iteratively checked its consistency with the next sentence, proceeding sequentially until the last sentence. After completing one round, we restarted the process from the first sentence again, repeating this cycle until all predicted results stabilized and no further changes were observed. At this point, the consistency checking process was terminated. By leveraging the contextual reasoning capabilities, this iterative approach enhanced the reliability and coherence of the coding process.  

\begin{figure}
\centering
\includegraphics[width=0.9\textwidth]{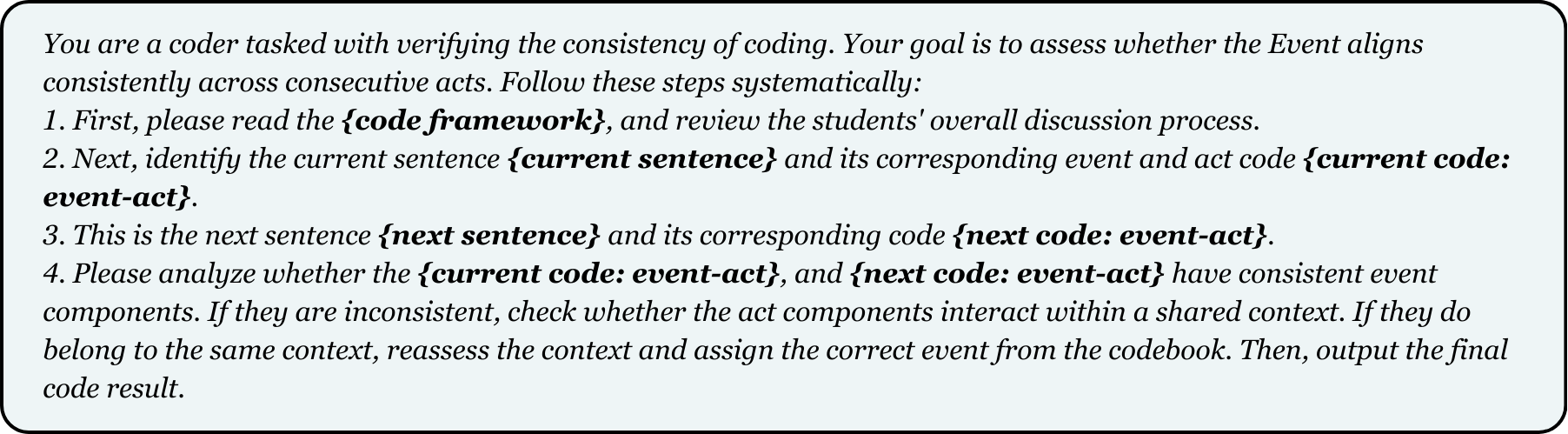}
\caption{\centering Prompt for Consistency Checking}\label{fig-cc}
\end{figure}

\subsubsection{Evaluation Metrics}

In our study, we employed the inner-annotator agreement (IAA) as a primary method for assessing the quality of coding~\cite{23-landis1977measurement}. IAA evaluates the clarity and interpretability of a codebook among coders, higher IAA ensuring that coders have a shared understanding of the coding framework, thereby enhancing the reliability of coded data. We employ Cohen’s k to measure the IAA, providing a quantitative measure of agreement between coders. Furthermore, we treat the coding results as a multi-class classification task. To comprehensively evaluate the performance of the coding process, we compute following metrics: Accuracy(ACC), macro and weighted F1 score(MF1, WF1), and IoU (MIoU, WIoU) for the predicted results. Macro metrics allow us to assess the average performance of each individual code~\cite{33-grandini2020metricsmulticlassclassificationoverview}, while weighted metrics provide an overall performance view(refer to Eq.s \ref{eq:weighted_f1} and \ref{eq:weighted_iou}). This dual evaluation approach enhances our ability to refine the coding process and improve the model's reliability in capturing the complexities of the data.

\begin{equation}\label{eq:weighted_f1}
\text{\textit{Weighted F1}} = \sum_{c \in C} \left( \text{\textit{F1}}_c \cdot \frac{\text{\textit{count}}(y_{\text{true}} = c)}{\text{\textit{total\_samples}}} \right),
\end{equation}

\begin{equation}\label{eq:weighted_iou}
\text{\textit{Weighted IoU}} = \sum_{c \in C} \left( \text{\textit{IoU}}_c \cdot \frac{\text{\textit{count}}(y_{\text{true}} = c)}{\text{\textit{total\_samples}}} \right).
\end{equation}

\section{Results}
\subsection{Separate Prompt vs Combined Prompt}
We compared the prediction results of combined and separate prompts for communicative events and communicative acts (refer to Table~\ref{fig-sc}). The results show that separate prompts outperform combined prompts across multiple metrics, including Cohen’s kappa, accuracy, F1, and IoU scores. This improvement may stem from decomposing complex contextual information, allowing more precise predictions. Notably, act predictions consistently achieve higher accuracy than event predictions, suggesting that smaller communicative units are easier to predict. From Table~\ref{fig-sc}, we can also see the prediction results of each individual LLM. Among the tested models, GPT-4-turbo achieved the highest kappa value (0.5919) for the event prompt, while GPT-4o reached the highest value for the act prompt (0.8632).

\begin{table}
\centering
\caption{Evaluation Metrics for Different Models Across Prompts on Validation Set}
\label{fig-sc}
\begin{tabular}{p{1.8cm} p{2.1cm} p{1.2cm} p{1.2cm} p{1.2cm} p{1.2cm} p{1.2cm} p{1.2cm}}
\hline
Prompt Type& Method& Kappa& ACC& MF1& MIoU& WF1& WIoU\\ 
\hline
\multirow{3}{*}{Combined}& gpt-4o & 0.5434 & 0.5569 & 0.3648 & 0.2836 & 0.5572 & 0.4251 \\
& gpt4-turbo & \textbf{0.5623} & 0.5608 & 0.3699 & 0.2886 & 0.5618 & 0.4293 \\
& deepseek-chat & 0.5228 & \textbf{0.5686} & \textbf{0.3726} & \textbf{0.2917} & \textbf{0.5711} & \textbf{0.4387} \\
\hline
\multirow{3}{*}{Event}& gpt-4o & 0.5369 & 0.6122 & 0.4351 & 0.3178 & 0.6182 & 0.4633 \\
& gpt4-turbo & \textbf{0.5919} & 0.6122 & \textbf{0.4523} & \textbf{0.3372} & 0.6190 & 0.4637 \\
& deepseek-chat & 0.5249 & \textbf{0.6204} & 0.4485 & 0.3346 & \textbf{0.6282} & \textbf{0.4733} \\
\hline
\multirow{3}{*}{Act}& gpt-4o & \underline{\textbf{0.8632}} & 0.8939 & 0.5602 & 0.4996 & 0.8908 & 0.8162 \\
& gpt4-turbo & \underline{0.8549} & 0.8939 & 0.5627 & 0.5021 & 0.8893 & 0.8131 \\
& deepseek-chat & \underline{0.8503} & \textbf{0.9020} & \textbf{0.5653} & \textbf{0.5071} & \textbf{0.8982} & \textbf{0.8284} \\
\hline
\end{tabular}
\end{table}

\subsection{Single LLM vs Multi-LLM prediction}
During the validation phase, we compared the performance of individual LLMs with the multi-LLM prediction results that combined these predictions into a collaborative decision task (refer to Eq.s~\ref{eq:freq}~and~\ref{eq:cfinal}). The results demonstrated that the multi-LLM approach consistently outperformed single-LLM predictions, achieving kappa values of 0.6353 and 0.8874 for events and acts, respectively (refer to Table~\ref{fig-cc}). This improvement can be attributed to the ensemble strategy, which mitigates the risk of over-relying on any single model's prediction. Each model functioned as a “weak model,” contributing predictions whose variability could be balanced out by others, leading to more stable and accurate classification.

\subsection{Result of Consistency Checking}  

\begin{figure}
    \centering
    \includegraphics[width=0.8\textwidth]{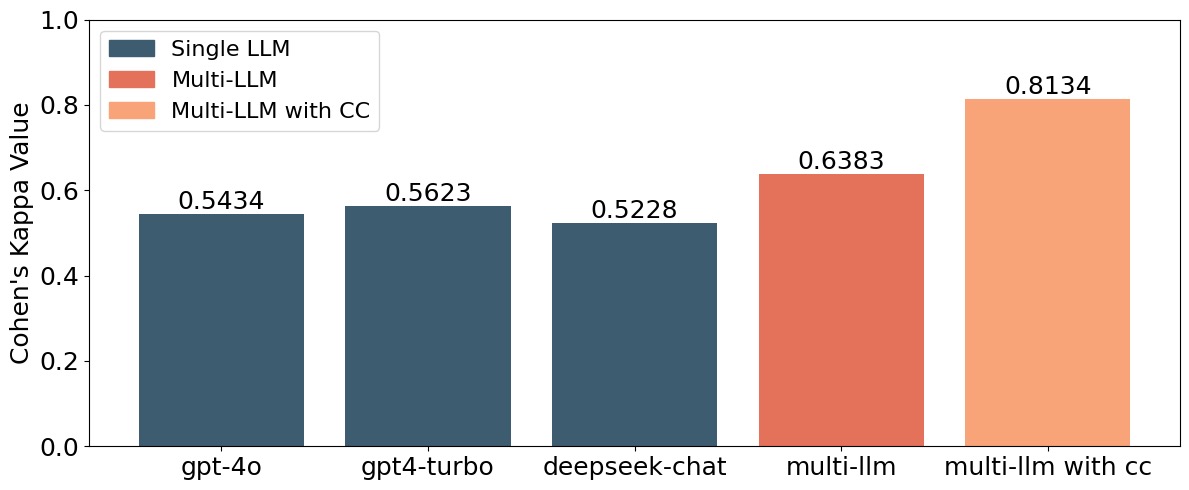}
    \caption{\centering Cohen's K Value of Final Coding Results on Validation Set}
    \label{fig-kappa}
\end{figure}

We conducted a comparative evaluation of prediction outcomes before and after implementing consistency checking (refer to Table~\ref{fig-cc}). The results show that consistency checking further enhanced prediction reliability, with the kappa value for events improving significantly from 0.6353 to 0.8267. This refinement also led to substantial gains in accuracy, F1 scores, and IoU metrics, demonstrating its effectiveness in reducing classification errors. Macro-F1 and IoU scores tend to be lower because they represent the average performance across all codes, making them sensitive to variations in accuracy across predicted codes. However, the observed improvement in these metrics suggests a more balanced overall performance across different categories.  Notably, 17\% of the coding results were corrected through this process, underscoring the role of consistency checking in optimizing overall predictive performance. From the kappa values of final coding results\footnote{Coding results refer to the annotation of an utterance after combining event and act.} combining event and act showed in Fig.~\ref{fig-kappa}, the multi-LLM approach (0.6383) outperforms single models, and consistency checking further boosts kappa to 0.8134, enhancing prediction reliability. 

\begin{table}
    \centering
    \caption{Evaluation Metrics for Multi-LLMs and Consistency Checking on Validation Set}
    \label{fig-cc}
    \begin{tabular}{p{1.5cm} p{3cm} p{1.2cm} p{1.2cm} p{1.2cm} p{1.2cm} p{1.2cm} p{1.2cm}}
        \hline
        Prompt Type & Method & Kappa & ACC & MF1 & MIoU & WF1 & WIoU \\ 
        \hline
        Event & Multi-LLMs & 0.6353 & 0.7061 & 0.5109 & 0.3896 & 0.7034 & 0.5518 \\
        & Multi-LLMs with CC & \underline{\textbf{0.8267}} & \textbf{0.8612} & \textbf{0.6825} & \textbf{0.5936} & \textbf{0.8606} & \textbf{0.7640} \\
        \hline
        Act& Multi-LLMs & \underline{0.8874} & 0.9184 & 0.6126 & 0.5503 & 0.9137 & 0.8562 \\
        & Multi-LLMs with CC & \underline{\textbf{0.9154}} & \textbf{0.9388} & \textbf{0.6505} & \textbf{0.6091} & \textbf{0.9319} & \textbf{0.8858} \\
        \hline
        Coding& Multi-LLMs & 0.6382 & 0.6588 & 0.4978 & 0.4145 & 0.6435 & 0.5224 \\
        Result& Multi-LLMs with CC & \underline{\textbf{0.8134}} & \textbf{0.7961} & \textbf{0.6766} & \textbf{0.6053} & \textbf{0.7900} & \textbf{0.6959} \\
        \hline
    \end{tabular}
\end{table}

\subsection{Performance Assessment on Test Set}

We evaluated the validated model on the test set, where performance remained strong(refer to Table~\ref{tab:kappa}).The slightly lower agreement in the test set compared to the validation set, as well as the reduced inter-annotator agreement, suggests that the test set may be inherently more challenging to annotate. The LLM's performance on the act prediction closely approximates human performance in both the validation and test sets, while its performance on the event prediction demonstrates stability and is also considered satisfactory. 

\begin{table}
    \centering
    \caption{Cohen's K Value for Validation and Test set}
    \label{tab:kappa}
    \begin{tabular}{l p{1.5cm} p{1.5cm} p{1.5cm} p{1.5cm}}
        \hline
        \multirow{2}{*}{Comparison} & \multicolumn{2}{c}{Validation Set} & \multicolumn{2}{c}{Test Set} \\
        \cline{2-3} \cline{4-5}
        & Event & Act & Event & Act\\
        \hline
        Inter-annotator Agreement (H1 vs H2) & 0.9898 & 0.9887 & 0.9175 & 0.9498 \\
        \hline
        Model-Human Agreement (M vs H1) & 0.8267 & 0.9209 & 0.7961 & 0.8703 \\
        \hline
        Model-Human Agreement (M vs H2) & 0.8267 & 0.9154 & 0.8444 & 0.8831 \\
        \hline
    \end{tabular}
\end{table}

\section{Discussion and Conclusion}

In this study, we enhanced LLMs' contextual understanding by leveraging the interdependence between communicative components, improving fully automated dialogue coding. Our key contributions are: 1) \textbf{Component-Based Prediction}: By segmenting utterances into event and act components, we boosted prediction accuracy by predicting them separately using role prompts and CoT. 2) \textbf{Multi-LLM Collaboration}: By using multiple LLMs to aggregate predictions, we improved robustness through ensemble methods. 3) \textbf{Consistency Checking}: By leveraging the interdependence between acts and events, we enhanced the prediction accuracy of events. The proposed LLM-Assisted Contextually Consistent Dialogue Coding Framework also provides a scalable solution for addressing contextual challenges in dialogue analysis. By focusing on structured communicative components, our approach offers valuable insights for educational research and practical applications in dialogue-based learning environments. A limitation of our current study is that it is based solely on the analysis of audio data, without incorporating other multimodal data. We will expand our analysis to provide a more comprehensive understanding of collaborative learning dynamics and enriching the insights derived from multiple sources of data.

%
%

\end{document}